\documentclass[runningheads]{llncs}

\usepackage{accv}

\usepackage{accvabbrv}          
\usepackage{graphicx}
\usepackage{booktabs}
\usepackage{multirow}
\usepackage{float}                   
\usepackage{algorithm}
\usepackage{algpseudocode}
\usepackage{placeins}                

\usepackage[accsupp]{axessibility}   

\usepackage{hyperref}

\usepackage{orcidlink}

\newcommand{\eps}{\epsilon}

\begin{document}

\title{Unlearnable Faces: Privacy Protection\\
Surviving Extraction Pipeline}
\titlerunning{Privacy Protection Surviving Extraction Pipeline}

\author{Byunghoon Oh \and
Sunghwan Park\orcidlink{0000-0002-0253-110X} \and
Jaewoo Lee\orcidlink{0000-0001-5887-2184}}

\authorrunning{B.~Oh et al.}

\institute{Chung-Ang University, Seoul, South Korea
\email{\{danny9807,tjdghks994,jaewoolee\}@cau.ac.kr}}

\maketitle

\begin{abstract}
Unlearnable examples keep publicly shared photos from being learned by unauthorized face-recognition models.
An imperceptible perturbation, added before sharing, makes any model trained on the protected photos fail on clean faces.
The perturbation is crafted on the shared image, however the attacker trains on the face it extracts, cropped and resized to the recognizer input, and under this extraction the protection collapses.
We propose LPID, which builds the extraction into the unlearnable-example objective.
LPID confines the perturbation to the extracted face region and optimizes it through a differentiable model of the extraction, concentrating its energy in the frequency band the extraction preserves.
Because this robustness is a property of the transform rather than of any identity, LPID
is re-optimized per album and protects even users it has never seen.
LPID attains the lowest attacker accuracy of all methods in every setting we evaluate, holding the attacker below $10\%$ under crop+resize extraction on identities unseen at protection time, while remaining imperceptible at $32.7$\,dB PSNR and $0.161$ LPIPS.


\end{abstract}

\section{Introduction}
\label{sec:intro}

\begin{figure}[t]\centering
\IfFileExists{figure1.pdf}{\includegraphics[width=\textwidth]{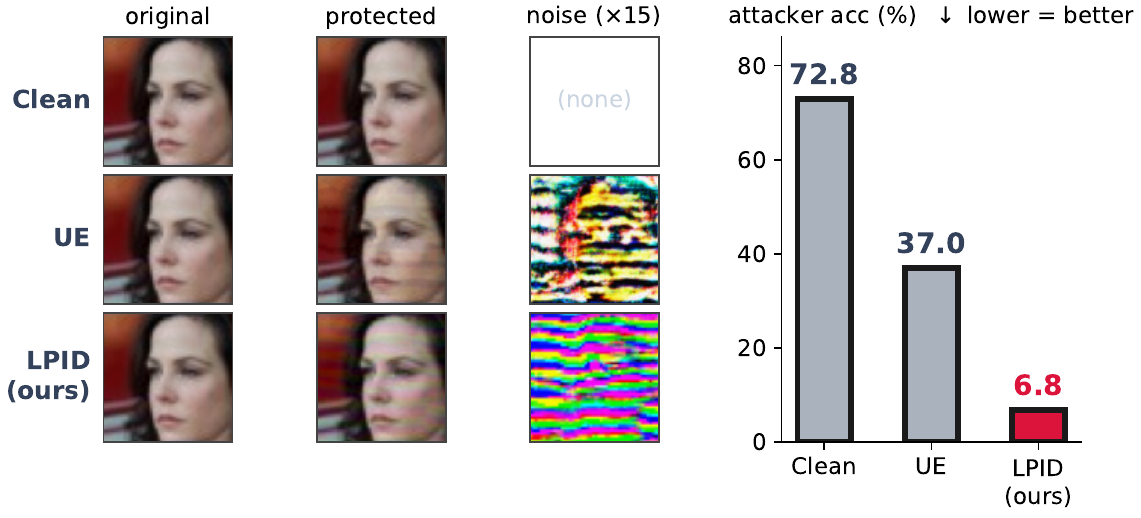}}{\fbox{\parbox[c][2.6cm][c]{0.98\textwidth}{\centering\emph{[Fig.~1 teaser: place figure1.pdf here]}}}}
\caption{The perturbation is imperceptible, yet only LPID's survives extraction. Rows (Clean, UE, LPID): the original face, the protected face, and the perturbation amplified $\times15$. UE's is high-frequency noise the resize attenuates; LPID's is low-frequency structure that survives. Right: attacker clean test accuracy after crop+resize to $112$ (Clean $72.8$, UE $37.0$, LPID $6.8$; \%, lower is better). Both are shown face-only so only the pipeline coupling differs, on identities unseen at protection time (\cref{tab:t1}).}
\label{fig:teaser}
\end{figure}

Unlearnable examples (UE) are a proactive, user-side defense that lets people keep their publicly shared photos from being learned by unauthorized face-recognition (FR) models. Such photos are scraped without consent to train models that can later identify their subjects, and a single company has already amassed billions of web-scraped faces to build one~\cite{guo2016msceleb}; modern face recognition is highly accurate~\cite{arcface} and remains effective with only a few images per identity~\cite{guo2017oneshot}, so a single public album is enough. To prevent this, before sharing the owner adds an imperceptible, error-minimizing perturbation that plants a spurious shortcut between image and label, so a model trained on the data learns the shortcut instead of identity and its accuracy drops to the random-guess level~\cite{ue}. This source-side guarantee (protect once, at upload time) has made UE an active line of privacy research~\cite{rem,sem,tue,lsp}.

UE have been strengthened along several methodological axes and carried over to new settings. Later work made the shortcut survive adversarial training~\cite{rem} and random perturbations~\cite{sem}, hold up under data augmentation~\cite{armor}, and resist spectral filtering~\cite{fuse}; for faces, the approach was extended to a generative protector, trained once, that lowers recognition accuracy~\cite{segue}. Across these advances the intended guarantee is the same: a face perturbed once, at upload time, remains unlearnable to any party that later collects it~\cite{ue,segue}.

However, in a realistic recognition pipeline this guarantee collapses (\cref{fig:teaser}), because the attacker does not train on the uploaded image itself. Shared photos place the face (often small, tens of pixels across~\cite{yang2016widerface,hu2017tinyfaces}) within a larger scene, and an attacker who scrapes one at scale~\cite{guo2016msceleb} first applies a transform before training: extracting the face and rescaling it to the model's input~\cite{arcface,mtcnn}.
This transform breaks the protection in two ways.
Spatially, the perturbation's budget is spread over the whole image, but the transform keeps only the face region and discards the rest.
Spectrally, whatever survives is high-frequency, and the rescaling acts as a low-pass filter that attenuates it.
Passed through this pipeline, existing perturbations retain almost none of their effect, and the attacker recovers the accuracy it would have without any protection (\cref{tab:t1}).

To address this, we propose \textbf{LPID} (Localized, Pipeline-coupled Identity Defense), which restores the protection by building the attacker's transform into the perturbation itself. LPID acts on the standard objective at the two points the transform attacks: it confines the perturbation to the face region the attacker extracts~\cite{salm}, and optimizes its generation through a differentiable model $T$ of that transform~\cite{lowkey,eot}. Optimizing through $T$ steers the perturbation's energy into the band the transform preserves, so the resulting error-minimizing perturbation reaches, and remains a learnable shortcut on, the very input the attacker trains on from scratch. Because this robustness is a property of the transform rather than of any particular face, LPID protects users unseen at protection time (identities held out from any training set, since the perturbation is re-optimized for each new album). Our main contributions are:

\begin{itemize}
    \item We propose LPID, a perturbation robust both to the attacker's transform and to users unseen at protection time.
    LPID confines the perturbation to the face region the attacker extracts and optimizes its generation through a differentiable model $T$ of that transform, the intervention point the standard objective had overlooked. It needs no knowledge of the protected user's identity; re-optimized per album, it applies unchanged to any unseen user (\cref{sec:method}).
    \item We establish why the protection works. A spectral analysis explains why existing perturbations are erased by the transform while ours survives (93\% of the perturbation energy in the resize-surviving band, \vs 33\% uncoupled), and a factorized ablation identifies which design choice is responsible for the protection (\cref{sec:spectral}).
    \item Extensive experiments show that LPID is effective in practice: across the attacker's transform, on identities unseen at protection time, and against several attacker architectures, LPID holds the attacker below $10\%$ across the crop+resize family, at $32.7$\,dB PSNR and $0.161$ LPIPS (\cref{sec:main-result,sec:transfer}).
\end{itemize}

\section{Related Work}

\subsection{Unlearnable Examples}
Unlearnable examples make training data useless to a model that trains on it, a form of availability poisoning~\cite{feng2019,fowl2021,sandoval2022} within the broader data-poisoning literature~\cite{shafahi2018}.
Like adversarial examples~\cite{szegedy2014,fgsm} they use small bounded perturbations, but to \emph{minimize} rather than maximize the training loss.
Huang \etal~\cite{ue} craft the original UE with a min-min bi-level objective that injects an error-minimizing perturbation, inducing a shortcut~\cite{geirhos2020} the model learns in place of real features; this perturbation is optimized at a single fixed resolution against a surrogate and is not robust to adversarial training~\cite{rem}.

To restore that robustness, REM~\cite{rem} adds an inner maximization (a costly min-min-max optimization), while SEM~\cite{sem} instead trains against random perturbations; both still operate on the whole image at a fixed resolution~\cite{rem,sem}.
TUE~\cite{tue} adds an optimizable class-wise separability term to make the perturbation transferable across datasets, and LSP~\cite{lsp} removes the surrogate entirely, synthesizing model-free, class-wise linearly separable perturbations, though with a larger, more visible perturbation.

For faces, Segue~\cite{segue} trains a class-conditional generator once to emit error-minimizing perturbations, with a distortion layer (adversarial training, blur, flip, sharpness, JPEG) for robustness; because it conditions on identity, it needs a label for each protected user and cannot cover users absent from its training set.

\subsection{Facial Privacy Protection}
Face-recognition systems can identify a person from a single photograph without their consent~\cite{arcface,facenet,cosface,sphereface}, and recognizers have already been built at scale from faces scraped off the web~\cite{guo2016msceleb}.
Facial privacy protection lets individuals defend their own images against this unauthorized recognition~\cite{fawkes,lowkey}; a complementary line instead replaces the face with a synthetic one via generative anonymization~\cite{deepprivacy,ciagan}, changing the published image rather than disrupting training.

Existing defenses differ mainly in \emph{when} the protective perturbation takes effect~\cite{fawkes,ue}.
\emph{Test-time} cloaks perturb an image so that an already-deployed recognizer misidentifies it~\cite{advmakeup,tipim}; such a cloak protects only that image, leaves the person's other photos usable, and does nothing to stop a recognizer from being trained on the collected data~\cite{fawkes,lowkey}.
\emph{Training-time} protections alter the images before they are collected, so that a model trained on them generalizes poorly to clean faces~\cite{ue,rem,segue}.

Across all of these, the perturbation is crafted and evaluated at a single fixed resolution, and prior work shows that simple input preprocessing (grayscale, JPEG, or resizing~\cite{iss,nonlinear}) already removes much of their protection.
We trace the resize case to the transform in the attacker's extraction pipeline and address it directly (\cref{sec:prelim}).
We take UE, REM, TUE, and LSP as our per-sample baselines and Segue as our generative baseline (\cref{sec:exp}).

\section{Preliminaries and Threat model}
\label{sec:prelim}

\subsection{Unlearnable examples: the standard objective}

Standard unlearnable examples assume the attacker trains on the protected image $x+\delta$ as is---the assumption our method overturns. 
They add an imperceptible, error-minimizing perturbation $\delta$ (bounded, $\|\delta\|_\infty\le\eps$) that injects a spurious shortcut between image and label, so a model trained on the protected data learns the shortcut in place of identity~\cite{ue}, by solving the min-min bi-level optimization problem
\begin{equation}
\min_\theta \; \min_{\|\delta\|_\infty \le \eps} \; \sum_i L\big(f_\theta(x_i + \delta_i),\, y_i\big),
\label{eq:ue}
\end{equation}
where $L$ is the cross-entropy loss and $f_\theta$ a surrogate recognizer, alternating an inner minimization that crafts $\delta$ against the current $f_\theta$ with an outer step that trains it.
Here $y_i$ is only the within-album grouping of images by identity (an arbitrary local index per identity), not membership in any pretrained label set; since the objective depends only on this local grouping and not on any fixed identity vocabulary, the same procedure runs unchanged on a new user's album, whose images simply receive fresh local indices.
Robust variants add an inner maximization (REM~\cite{rem}) or train against random noise (SEM~\cite{sem}) to survive adversarial training, but all inherit this same assumption.

\subsection{Threat model}

\noindent\textbf{Attacker.}
The attacker collects the published photos and trains its own face recognizer from scratch, holding no pre-trained model~\cite{ue,segue}.
It scrapes the images~\cite{guo2016msceleb} (possibly already recompressed by the hosting platform at upload) and applies the preprocessing every modern face recognizer requires~\cite{mtcnn,arcface}.
It detects and aligns the face~\cite{mtcnn,retinaface}, crops it out, and resizes the crop to the recognizer's fixed input resolution $s$ (\eg $112$ for ArcFace~\cite{arcface}), optionally followed by Gaussian blur~\cite{segue}, before training a classifier.
We call this detect--crop--resize sequence the attacker's \emph{extraction pipeline}, and write C+R$s$ for its crop-and-resize core.
Crucially, detection and alignment merely locate the face; the crop-and-resize is the step that actually reshapes the pixels carrying the perturbation, and it is unavoidable, since every recognizer needs a fixed input resolution~\cite{arcface}.
This is why our method models exactly this crop-and-resize as a differentiable transform $T$ and couples the perturbation to it.

We score protection by the attacker's model, an identity classifier over the collected people (25 identities here).
Its clean test accuracy is the fraction of held-out photos it labels with the correct person, where lower means better protection.
Protection lowers this accuracy rather than hiding the face, which stays fully visible; random-guess accuracy is $\approx4\%$. Unlike the aligned, full-frame faces on which recognizers are usually benchmarked~\cite{lfw,knoche2021xqlfw}, the scraped photos here hold a small face within a larger scene~\cite{yang2016widerface,hu2017tinyfaces}.

\noindent\textbf{Defender.} The defender must make a face unlearnable before publishing it, without knowing how the attacker will later process the image. Given a face of native size $s_0$ at a known box $(x_0,y_0)$ with label $y$ in the published image, the defender adds the perturbation $\delta$ at budget $\eps=8/255$, confined by a mask $m$ to the face box and crafted against a surrogate $f_\theta$ (a ResNet-18~\cite{resnet}) that shares no weights with the attacker's network; the perturbation is committed once, before any recognizer has seen it.

Because it is committed blindly, the protection must survive two unknowns.
First, the defender does not know the attacker's input size $s$, resize kernel, or the platform's JPEG quality (each fixed after publication and never disclosed), so protection must hold across the whole family of plausible resize operators, not one known target (\cref{sec:spectral} shows why a single generation target suffices).
Second, it must protect \emph{unseen} users, whose identities are disjoint from the protector's training set and are therefore unknown when the protection is built, an \emph{identity-disjoint} (subject-independent) protocol~\cite{fawkes,lowkey,gunther2017openset}; our per-sample perturbation is re-optimized on each new album, so an unseen identity is handled exactly as a seen one, whereas a generative protector conditioned on an identity label cannot cover a user it was never trained for.

\begin{figure}[!t]\centering
\IfFileExists{figure2.pdf}{\includegraphics[width=\textwidth]{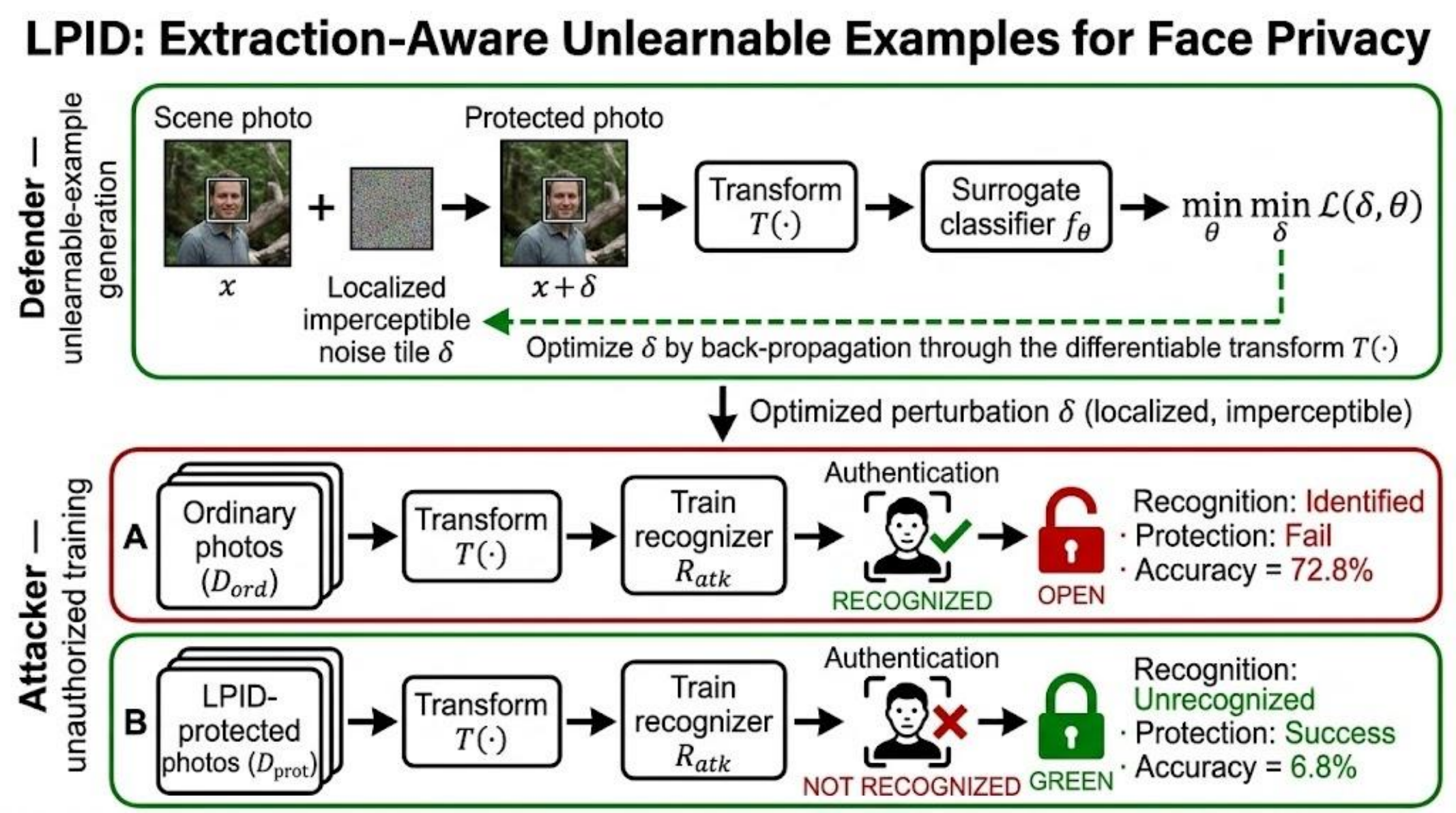}}{\fbox{\parbox[c][3cm][c]{0.98\textwidth}{\centering\emph{[Fig.~2 method overview: place figure2.pdf here]}}}}
\caption{LPID overview. \textbf{Defender.} LPID crafts a localized, imperceptible perturbation $\delta$ by solving the unlearnable-example min-min objective \emph{through} the attacker's differentiable transform $T(\cdot)$: $\delta$ is added to the scene photo $x$, the composite $x+\delta$ is passed through $T$ into a surrogate classifier $f_\theta$, and the error-minimizing loss $\mathcal{L}(\delta,\theta)$ is back-propagated through $T$ to update $\delta$; the optimized $\delta$ yields the published protected photo $x+\delta$. \textbf{Attacker.} The attacker applies the same $T$ and trains its own face recognizer $R_{\mathrm{atk}}$ from scratch, an identity classifier over the protected identities, distinct from the defender's surrogate $f_\theta$. Trained on ordinary photos ($D_{\mathrm{ord}}$), $R_{\mathrm{atk}}$ recognizes the subject, so protection fails ($72.8\%$ clean test accuracy, open lock); trained on LPID-protected photos ($D_{\mathrm{prot}}$), it cannot, so protection succeeds ($6.8\%$, closed lock; C+R112, \cref{tab:t1}).}
\label{fig:method}
\end{figure}

\section{LPID: Localized, Pipeline-coupled Identity Defense}
\label{sec:method}

\subsection{Extraction-aware noise generation}

We build the attacker's extraction pipeline into the standard UE objective. As \cref{sec:prelim} describes, the attacker does not train on the protected image $x+\delta$ but on the face it extracts from it; we model that extraction as a differentiable operator $T$ (\cref{fig:method}; defined below) and craft the perturbation \emph{through} it. In the standard objective \cref{eq:ue}, $x+\delta$ then becomes $T(x+m\odot\delta)$, giving the LPID objective
\begin{equation}
\min_\theta \; \min_{\|\delta\|_\infty \le \eps} \; \sum_i L\big(f_\theta( T(x_i + m_i \odot \delta_i) ),\, y_i\big).
\label{eq:lpid}
\end{equation}
This adds two ingredients to \cref{eq:ue}, one per mismatch.

\emph{Localization:} $m$ is a binary mask (one on the face box the attacker crops, zero elsewhere), so $m\odot\delta$ confines the perturbation to the region the extraction keeps, following the region-masking of~\cite{salm}.

\emph{Coupling:} rather than crafting $\delta$ in pixel space, we optimize it through the differentiable operator $T$~\cite{lowkey,eot}, so the gradient that shapes $\delta$ is the one the attacker's own pipeline will pass.
We instantiate $T$ as a differentiable model of the extraction, a crop of the face box followed by a bilinear resize to the surrogate's input resolution.
The attacker's own output resolution is unknown to the defender, and \cref{sec:spectral} shows why generating at a single fixed resolution suffices.
Given the box at top-left $(x_0,y_0)$ with native side $s_0$, $T$ composes the crop (a $0/1$ selection of the box) with the resize:
\begin{equation}
T(z) = R_{s_0 \rightarrow 224}\big(\, z[\,:,\, y_0{:}y_0{+}s_0,\; x_0{:}x_0{+}s_0\,]\,\big),
\label{eq:T}
\end{equation}
with \texttt{align\_corners}$=$false, where output pixel $o$ reads input coordinate $u(o)=(o+\tfrac12)\, s_0/224-\tfrac12$ and interpolates its two neighbours.
Both operations are linear in the pixels, so $T = R\,S_{\text{box}}$ is a fixed linear map (differentiable, four-tap-sparse Jacobian).
The crop $S_{\text{box}}$ confines the gradient to the face box, and the resize $R$ means the surrogate receives $R\delta$, so the inner minimization concentrates the perturbation energy in the band the resize preserves; \cref{sec:spectral} makes this precise.

\subsection{Why LPID survives the transform}
\label{sec:spectral}

\begin{figure}[t]\centering
\IfFileExists{figure3.pdf}{\includegraphics[width=\textwidth]{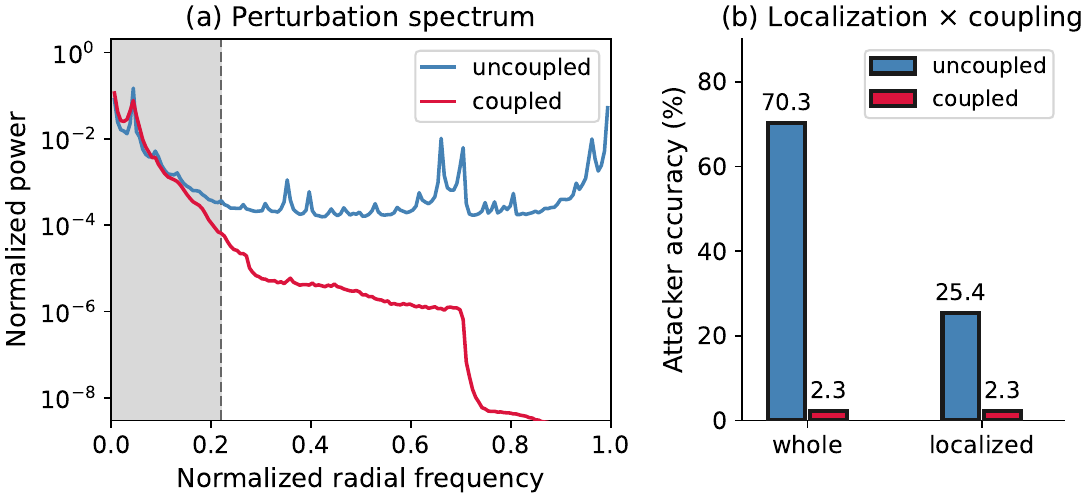}}{\fbox{\parbox[c][3.6cm][c]{0.98\textwidth}{\centering\emph{[Fig.~3: combined spectrum (a) + ablation bar chart (b); place figure3.pdf here]}}}}
\caption{The coupling mechanism. \textbf{(a)} Radially-averaged perturbation spectrum, coupled (red) \vs uncoupled (blue). The dashed line marks the native Nyquist cutoff $s_0/2$ on the normalized axis; the shaded region below it is the resize-surviving band. Coupling concentrates the energy there ($\rho_{\mathrm{LF}}=0.93$ \vs $0.33$ uncoupled; \cref{eq:rholf}), whereas an uncoupled perturbation spreads it across the full band. \textbf{(b)} Factorized ablation (localization $\times$ coupling; 50 unseen identities, C+R112, attacker accuracy \%, lower is better): localization alone is insufficient ($25.4$), coupling is decisive ($2.3$).}
\label{fig:mechanism}
\end{figure}

\emph{Coupling} is what makes $\delta$ survive.
It concentrates the perturbation's energy in the frequency band the resize keeps.
We make this precise with standard sampling theory~\cite{guo,sharma} (introducing no new theory), then confirm it with a factorized ablation (\cref{fig:mechanism}).

Because $T$ is linear, $T(x+\delta)=Tx+T\delta$ (\cref{eq:T}), and the surrogate only ever receives $T\delta$. 
A resize is a low-pass filter whose cutoff is half the smaller of its input and output resolutions~\cite{guo,sharma}; it transmits the band below that cutoff and attenuates the rest, echoing the frequency perspective on CNN generalization and robustness~\cite{wang2020hf,yin2019}.
Let $V$ be the face-box subspace band-limited to the native rate $s_0/2$, the largest band any resize of the face can keep.
Optimizing $\delta$ through $T$ gives the inner loop no gradient toward the attenuated high-frequency components, so $\delta$'s energy accumulates in $V$; an uncoupled, pixel-space perturbation instead spreads its budget across the full band, most of which the resize discards. 

We quantify the concentration by the surviving-band energy fraction, with $P_V$ the ideal low-pass projection onto $V$:
\begin{equation}
\rho_{\mathrm{LF}} := \frac{\|P_V\,\delta\|^2}{\|\delta\|^2}.
\label{eq:rholf}
\end{equation}
Coupling raises this fraction from $\rho_{\mathrm{LF}}=0.33$ (uncoupled) to $0.93$ (coupled) at the native cutoff (\cref{fig:mechanism}a), and the crop+resize columns of \cref{tab:t1} bear this out against the baselines.

This also explains why the defender can generate at a single resolution without knowing the attacker's.
Energy placed below the native cutoff $s_0/2$ is kept by any resize that preserves that band (in particular by any target at least the native size, which merely up-samples), so protection does not require matching the attacker's resolution.
\Cref{tab:t1} confirms this across the full range of native face sizes and attacker resolutions we evaluate (\cref{sec:setup}).

A factorized ablation confirms that coupling, not localization, is responsible (\cref{fig:mechanism}b).
We toggle localization (whole image \vs the face box) and coupling (crafted in pixel space \vs through $T$) on our own perturbation and measure the attacker's clean test accuracy.
Whole-image, uncoupled noise gives no protection (attacker $70.3$, near the clean $71.1$); adding localization alone recovers only part ($25.4$); adding coupling drives the attacker to $2.3$, near random. Coupling is the decisive factor, and localized$+$coupled equals whole$+$coupled because $T$'s crop already zeros the off-box gradient.

\subsection{Algorithm}
\label{sec:algo}

\Cref{alg:lpid} solves \cref{eq:lpid} with the standard alternating min-min optimization of unlearnable examples~\cite{ue}, adding only the operator $T$ to each step.
Per minibatch, an inner projected-gradient (PGD)~\cite{madry} step updates the perturbation $\delta$ (a sign-gradient step taken through $T$, then clamped to the $\eps$-ball and masked to the face box), and an outer stochastic-gradient (SGD)~\cite{bottou} step trains the surrogate $f_\theta$.
The clamp is applied before $T$ so that $T$ stays the linear operator of \cref{eq:T}.
We use the surrogate and budget of \cref{sec:prelim} (ResNet-18, $\eps=8/255$) with $K$ inner PGD steps per outer step; $K$, the step sizes, and the epoch budget $E$ are given in \cref{sec:setup}.
As \cref{sec:spectral} shows, the surviving band is fixed by the native face size, not by the generation resolution.

\begin{algorithm}[H]
\caption{LPID: extraction-aware UE generation}
\label{alg:lpid}
\begin{algorithmic}[1]
\Require images $\{x_i\}$, masks $\{m_i\}$, labels $\{y_i\}$; budget $\eps$; inner steps $K$, step $\alpha$; outer learning rate $\eta$; surrogate $f_\theta$; epochs $E$
\State $\delta_i \gets 0$ for all $i$
\For{epoch $=1..E$}
  \For{minibatch $(x,m,y)$}
    \For{$k=1..K$}
      \State $\tilde x \gets \mathrm{clip}_{[0,1]}(x + m\odot\delta)$ \Comment{clamp before $T$}
      \State $z \gets T(\tilde x)$ \Comment{crop box, bilinear resize}
      \State $\delta \gets \mathrm{clip}_{[-\eps,\eps]}(\delta - \alpha\,\mathrm{sign}(\nabla_\delta L(f_\theta(z),y))) \odot m$
    \EndFor
    \State $\theta \gets \theta - \eta\, \nabla_\theta L(f_\theta(T(\mathrm{clip}(x+m\odot\delta))), y)$
  \EndFor
\EndFor
\State \Return $\{m_i\odot\delta_i\}$
\end{algorithmic}
\end{algorithm}

\FloatBarrier
\section{Experiments}
\label{sec:exp}

\begin{table}[t]
\centering
\footnotesize
\caption{Main result: attacker clean test accuracy (\%, lower is better) under the realistic small-face extraction pipeline; 25 unseen identities, random-guess $\approx$4\%, 3 seeds, lowest per column bold. Per-sample UE are shown both whole-image and localized face-only (matching LPID's coverage). Columns: crop+resize to $224$/$112$ (C+R$s$); a bicubic variant; quality-$85$ JPEG before C+R112; and $+$blur.}
\begin{tabular}{llccccc}
\toprule
Method & & C+R224 & C+R112 & C+R112 (bic.) & JPEG85 & C+R112 +blur \\
\midrule
Clean (no prot.) & & 75.3 & 72.8 & 72.0 & 72.1 & 72.3 \\
\midrule
\multirow{2}{*}{UE}  & whole     & 70.1 & 70.1 & 71.4 & 70.3 & 69.4 \\
                     & face-only & 24.4 & 37.0 & 38.2 & 39.4 & 37.7 \\
\midrule
\multirow{2}{*}{REM} & whole     & 74.8 & 74.0 & 74.0 & 72.8 & 72.6 \\
                     & face-only & 36.9 & 74.0 & 73.3 & 73.2 & 72.6 \\
\midrule
\multirow{2}{*}{TUE} & whole     & 75.0 & 72.1 & 74.0 & 73.1 & 72.9 \\
                     & face-only & 74.3 & 73.5 & 73.8 & 72.5 & 71.6 \\
\midrule
\multirow{2}{*}{LSP} & whole     & 74.3 & 72.7 & 72.8 & 72.7 & 72.7 \\
                     & face-only & 74.5 & 73.3 & 72.0 & 73.6 & 71.7 \\
\midrule
Segue                & & 72.1 & 72.1 & 72.4 & 71.5 & 71.9 \\
\midrule
\textbf{LPID (ours)} & & \textbf{4.9} & \textbf{6.8} & \textbf{6.3} & \textbf{20.8} & \textbf{6.9} \\
\bottomrule
\end{tabular}
\label{tab:t1}
\end{table}

\subsection{Setup}
\label{sec:setup}

\noindent\textbf{Datasets.} We build composites by pasting CASIA-WebFace~\cite{casiawebface} faces at native sizes $\{50,75,100,150,200\}$ (small to near-frame-filling) at random positions on Places365~\cite{places365} backgrounds on a $224\times224$ canvas; the smaller sizes match the small-face regime of scraped imagery~\cite{yang2016widerface,hu2017tinyfaces}. We use an 80/20 train/test split per identity.

\noindent\textbf{Baselines.} We re-implement all baselines, UE~\cite{ue}, REM~\cite{rem}, TUE~\cite{tue}, LSP~\cite{lsp}, and Segue~\cite{segue}, from their papers and, where available, official code, with the reported hyperparameters and no added resolution augmentation. All $\ell_\infty$ methods use $\eps=8/255$ and a ResNet-18 surrogate; LSP uses its native $\ell_2$ budget and is model-free. Segue's generator is conditioned on a training identity (\cref{sec:prelim}); since in a real deployment the users to protect are not known when the protector is built, we query it with a nearest-identity pseudo-label in this realistic unknown-user setting (\cref{tab:t1}), and grant it the ground-truth label only where a control fixes a known user (\cref{sec:aligned,sec:fullres-control}).

\noindent\textbf{Extraction.} We recall C+R$s$ from \cref{sec:prelim}, a crop of the face box then a resize to $s\times s$ (bic.\ $=$ bicubic kernel). We evaluate the endpoints $s=112,224$ of the $\{112,160,224\}$ recognizer family. JPEG85$\rightarrow$C+R112 applies q85 compression before extraction (platform recompression); $+$blur adds Gaussian blur after the resize.

\noindent\textbf{Metric.} The metric is the attacker's clean test accuracy (\%, lower is better; \cref{sec:prelim}), with the lowest per column in bold. All results are on \emph{unseen} identities (\cref{sec:prelim}), averaged over 3 seeds; each table and figure states its identity count (random-guess $\approx$4\% for 25, $\approx$2\% for 50).

\noindent\textbf{Implementation details.} LPID crafts $\delta$ for $E=60$ epochs with $K=20$ inner PGD steps of size $\alpha=0.8/255$, through a bilinear $T$ at target $224$. The ResNet-18 surrogate and the attacker are trained from scratch with SGD at learning rate $0.025$; the attacker uses a $30$-epoch schedule, extended to a fixed $80$ epochs for the cross-architecture study (\cref{tab:t3}).

\subsection{Main result: existing UE break under the extraction pipeline}
\label{sec:main-result}

\begin{figure}[t]\centering
\IfFileExists{figure4.pdf}{\includegraphics[width=\textwidth]{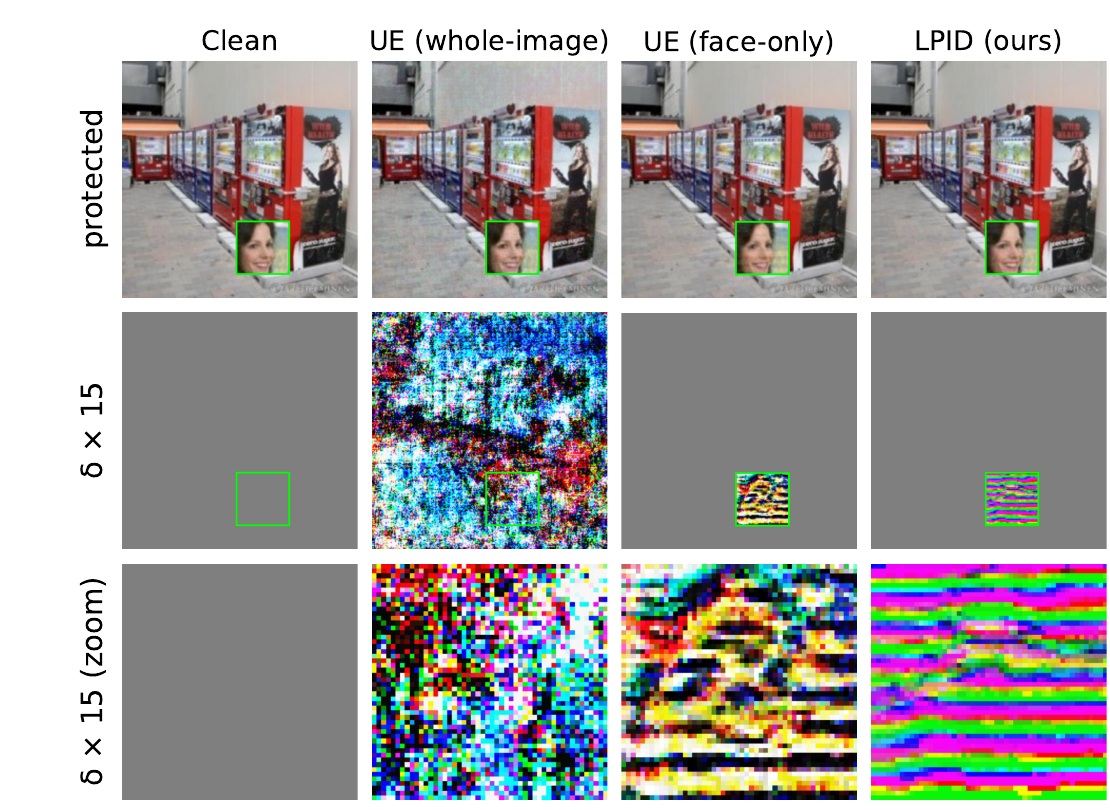}}{\fbox{\parbox[c][3.4cm][c]{0.98\textwidth}{\centering\emph{[Fig.: whole \vs face-only \vs LPID contrast; place figure4.pdf here]}}}}
\caption{Where each method places its perturbation, exposing the spatial mismatch that neutralizes whole-image UE. Columns: Clean, whole-image UE, face-only UE, and LPID; rows: the protected composite, the perturbation ($\times15$), and a face-region zoom ($\times15$). Whole-image UE spreads its budget across the scene, so the attacker's crop discards most of it; face-only and LPID sit inside the kept box, where UE's residual is high-frequency grain the resize attenuates while LPID's is low-frequency structure in the surviving band (the visual counterpart of \cref{tab:t1}'s whole \vs face-only rows).}
\label{fig:contrast}
\end{figure}

\Cref{tab:t1} reports every method under the realistic small-face extraction pipeline. Applied natively over the whole published image, the per-sample UE give essentially no protection.
The attacker's crop discards the budget outside the face box, leaving the attacker at 69--75\%, near the 72--75\% clean accuracy (the spatial mismatch; \cref{fig:contrast}).
Localizing them to the face box the attacker crops, matching LPID's spatial coverage (the face-only rows), does not rescue them: at C+R224 only the error-minimizing pair improves appreciably (UE 70.1 to 24.4, REM 74.8 to 36.9), and at the recognizer's 112 input the resize attenuates the high-frequency content of every localized baseline, so the attacker still reaches 37--74\% (UE 37.0, REM/TUE/LSP 73--74). Localization removes the spatial mismatch but not the spectral one, which coupling addresses, as the ablation in \cref{sec:spectral} shows.

For these unseen users, Segue has no valid identity to condition on and gives essentially no protection ($\sim$72\%, \cref{sec:setup}).
LPID, a per-sample method localized to the crop box and coupled to the differentiable transform, holds the attacker below $7\%$ across the crop+resize family (4.9--6.9\%), the sole lowest method in every crop+resize column. For UE and REM, which protect strongly when no extraction is applied (\cref{sec:fullres-control}), this collapse is attributable to extraction itself; TUE and LSP do not protect even at full resolution, so extraction is not what causes their failure. The upload-JPEG column is a separate axis.
Recompression at upload (q$\approx$70--85~\cite{chuman}) attenuates even the resize-surviving band, degrading LPID to $20.8\%$ at q85, a limitation it shares with all $\eps$-bounded perturbations~\cite{iss,nonlinear}.

\subsection{Imperceptibility}
\label{sec:imperceptibility}

\begin{figure}[tp]\centering
\IfFileExists{figure5.pdf}{\includegraphics[width=\textwidth]{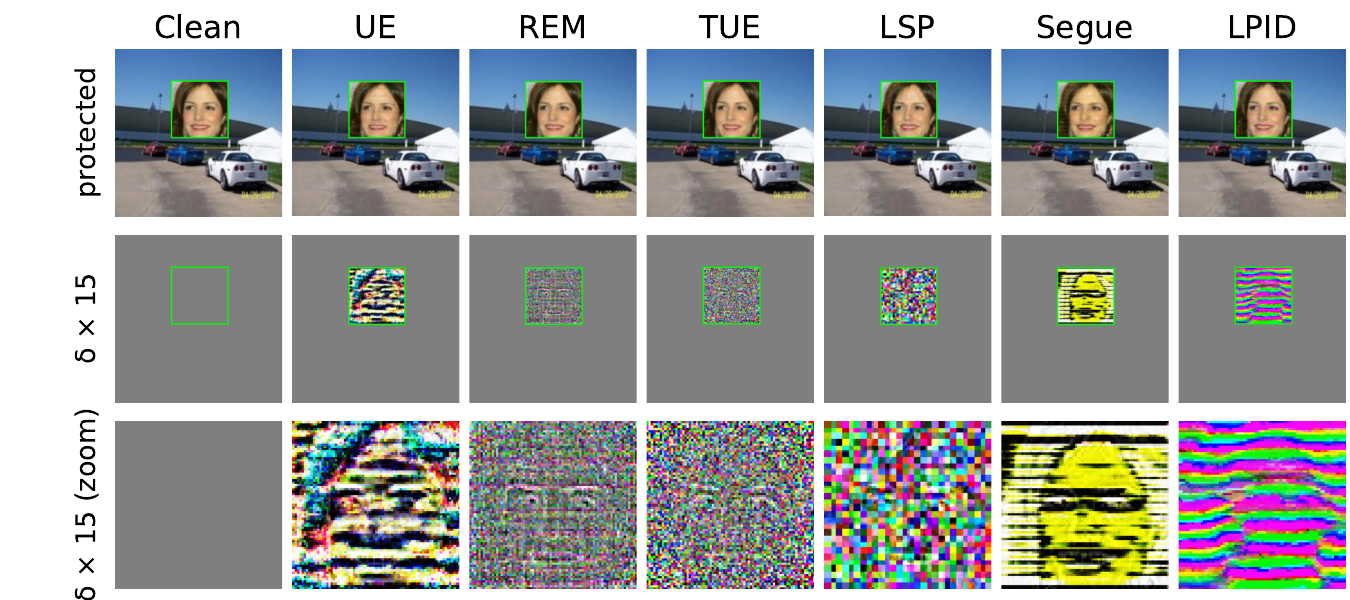}}{\fbox{\parbox[c][2.1cm][c]{0.98\textwidth}{\centering\emph{[Fig.: perturbation textures across methods; place figure5.pdf here]}}}}
\caption{Perturbation textures. Rows: the protected composite, the perturbation ($\times15$), and a face-region zoom ($\times15$); all are imperceptible at native resolution (top). The per-sample UE and LSP are high-frequency grain the resize attenuates, and Segue a structured generator pattern, whereas LPID is low-frequency structure concentrated in the resize-surviving band---the only texture shaped to survive the crop+resize (\cref{sec:spectral}).}
\label{fig:textures}
\end{figure}

\Cref{tab:t4} reports PSNR/SSIM/LPIPS on the protected faces to control for a possible confound in perturbation budget; PSNR (dB) and SSIM~\cite{ssim} measure pixel and structural fidelity to the clean face (higher is better), while LPIPS~\cite{lpips} is a learned perceptual distance (lower is better). LPID (32.7\,dB, 0.917, 0.161) is comparable to UE on PSNR and SSIM and less perceptible than Segue on LPIPS (0.161 \vs 0.240). LPID's protection does not rely on a larger perturbation. Like UE/REM/TUE/Segue it respects $\eps=8/255$, whereas LSP's $\ell_2$ budget yields a larger residual (max$|\delta|=43.7/255$). \Cref{fig:textures} shows the corresponding textures. All are imperceptible at native resolution, but the per-sample UE and LSP are high-frequency grain while LPID is the low-frequency structure the resize preserves.

\begin{table}[t]\centering
\footnotesize
\caption{Imperceptibility of the perturbation on the protected faces (CASIA-WebFace on Places365 composites; $\eps=8/255$ for all $\ell_\infty$ methods, LSP uses its native $\ell_2$ budget). Arrows give the better direction.}
\begin{tabular}{lcccc}
\toprule
Method & PSNR $\uparrow$ & SSIM $\uparrow$ & LPIPS $\downarrow$ & max$|\delta|$ (/255) \\
\midrule
UE   & 32.9 & 0.912 & 0.104 & 8.0 \\
REM  & 39.3 & 0.953 & 0.073 & 8.0 \\
TUE  & 36.5 & 0.907 & 0.165 & 8.0 \\
LSP  & 31.3 & 0.867 & 0.252 & 43.7 \\
Segue & 31.3 & 0.900 & 0.240 & 8.0 \\
\textbf{LPID (ours)} & 32.7 & 0.917 & 0.161 & 8.0 \\
\bottomrule
\end{tabular}
\label{tab:t4}
\end{table}

\subsection{Generalization to aligned faces}
\label{sec:aligned}

To rule out that LPID's protection is an artifact of our small-face-in-scene composites, we evaluate it in the standard aligned-face setting with no compositing: raw CASIA-WebFace faces at $224$ (25 unseen identities), with the attacker resizing to the $112$ recognizer input. LPID limits the attacker to 6.3\%, compared with 35.7\% for UE and 21.2\% for Segue given its ground-truth label (clean 75.3\%). The protection therefore transfers to real aligned faces and is not a property of the composite construction. In both settings the background is discarded by the attacker's crop and is absent from the extracted face; protection is determined only by the perturbation's survival through the resize.

\subsection{Transferability}
\label{sec:transfer}

\Cref{tab:t3} evaluates each method's perturbation against four attacker architectures (ResNet-18~\cite{resnet}, MobileNet-V2~\cite{mobilenetv2}, DenseNet-121~\cite{densenet}, VGG-16~\cite{vgg}) trained from scratch at C+R112, testing whether LPID's protection depends on the ResNet-18 surrogate it was crafted on. LPID limits the attacker to 4.4--9.1\% accuracy across all four architectures, the lowest of every method and far below the per-sample baselines (33.2--76.3\%); its protection is therefore not specific to the surrogate. Segue is omitted here; it is evaluated in \cref{tab:t1}.

\begin{table}[t]\centering
\footnotesize
\caption{Transferability: attacker clean test accuracy (\%, lower is better) at C+R112 across four backbones trained from scratch (80 epochs, 25 unseen identities; lowest per column bold). The LPID perturbation is generated on a ResNet-18 surrogate and evaluated unchanged against each backbone.}
\begin{tabular}{lcccc}
\toprule
Method & ResNet-18 & MobileNet-V2 & DenseNet-121 & VGG-16 \\
\midrule
Clean & 68.8 & 65.1 & 64.7 & 75.2 \\
UE   & 38.0 & 33.2 & 43.1 & 37.5 \\
REM  & 67.9 & 63.7 & 63.9 & 76.3 \\
TUE  & 69.2 & 68.9 & 66.1 & 75.1 \\
LSP  & 68.5 & 66.4 & 62.3 & 75.3 \\
\textbf{LPID (ours)} & \textbf{6.3} & \textbf{4.4} & \textbf{9.1} & \textbf{6.9} \\
\bottomrule
\end{tabular}
\label{tab:t3}
\end{table}

\begin{figure}[t]\centering
\IfFileExists{figure6.pdf}{\includegraphics[width=\textwidth]{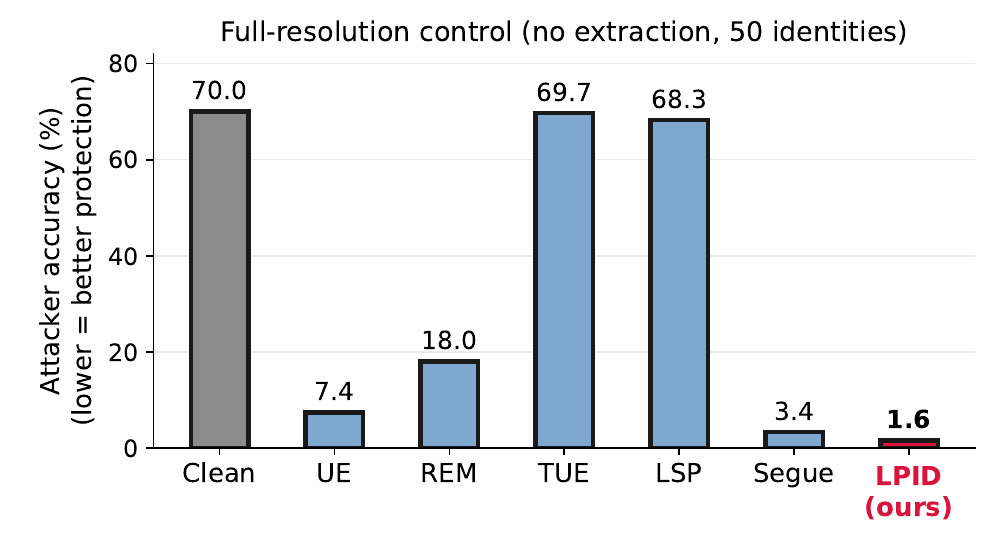}}{\fbox{\parbox[c][3.2cm][c]{0.95\textwidth}{\centering\emph{[Fig.~6 full-resolution control bar chart: place figure6.pdf here]}}}}
\caption{Full-resolution control (no extraction, 50 unseen identities). Attacker clean test accuracy (\%, lower is better; Clean $70.0$). UE and REM protect strongly here ($7.4$, $18.0$) yet collapse under extraction (\cref{tab:t1}), confirming that extraction, not a weak reimplementation, causes the collapse; TUE and LSP do not protect even here ($69.7$, $68.3$); Segue (true label) and LPID protect throughout ($3.4$, $1.6$).}
\label{fig:fullres}
\end{figure}

\subsection{Full-resolution control}
\label{sec:fullres-control}

To confirm the collapse of \cref{tab:t1} is caused by extraction rather than weak baselines, \cref{fig:fullres} reports the attacker's clean test accuracy on full-resolution faces (224, no crop+resize), the standard UE setting (50 unseen identities). UE and REM protect strongly here (7.4 and 18.0, \vs clean 70.0), so our reimplementations are faithful and their collapse to 69--75\% under extraction (\cref{tab:t1}) is attributable to the pipeline. TUE and LSP do not protect even at full resolution (69.7, 68.3), so we restrict the extraction-causes-the-collapse argument to UE and REM. Given the true label, Segue also protects here (3.4), matching LPID (1.6); the realistic case, where that label is unavailable, is \cref{tab:t1}.

\FloatBarrier
\section{Conclusion}

Unlearnable examples are crafted in pixel space, assuming the perturbation reaches the model unchanged, but a face-recognition attacker first applies a transform that existing UE do not model: it crops the face and resizes it, so the crop discards the perturbation outside the box and the resize attenuates what remains.
We showed that existing per-sample UE collapse under this pipeline and proposed LPID, which generates extraction-aware noise, localized to the face the attacker crops and optimized through the crop+resize, so that, by construction, the optimization concentrates the surviving energy in the resize-surviving band.
On identities unseen at protection time, LPID attains the lowest attacker accuracy of all methods across the crop+resize family, showing that the protection holds in the realistic deployment where the users to protect are unknown in advance. 
These results indicate that face-privacy perturbations should be evaluated under the attacker's full extraction pipeline, and that coupling the perturbation to the extraction operator, rather than to a fixed set of distortions, is what lets the protection survive it. 
We hope this extraction-aware perspective offers a practical step toward facial privacy protection in the real world.

\newpage
\bibliographystyle{splncs04}
\bibliography{main}
\end{document}